**Title:** Cuttlebot: a platform demonstration for complex, autonomous, bio-inspired swimmers

**Authors:**

A. N. White,[1] A. L. Li,[2,3] A. H. Yin,[1] D. Roseman,[4] V. Saro-Cortes,[4] H. Wiswell,[4] A. Wissa,[4] M. Duduta[1,3]*

**Affiliations:**

[1]School of Mechanical, Aerospace, and Manufacturing Engineering, University of Connecticut; Storrs, CT, U.S.A.
[2]Department of Mechanical and Industrial Engineering, University of Toronto; Toronto, Ontario, Canada.
[3]University of Connecticut, Institute of Materials Science; Storrs, CT, U.S.A.
[4]Department of Mechanical and Aerospace Engineering, Princeton University; Princeton, NJ, U.S.A.

*mihai.duduta@uconn.edu

**Abstract:** Increasing interest in deep-sea operations and resources motivates the development of ecologically sensitive but environmentally durable robots. Dielectric elastomer actuator artificial muscles are good candidates for powering such systems due to their pressure and temperature tolerance and soft makeup, but they are difficult to integrate with robotic systems. This work presents an autonomous robotic platform: the CORE, capable of driving six artificial muscles while sensing visual and spatial information. To validate the platform, we developed the Cuttlebot – a cuttlefish-inspired robot that swims in three dimensions using undulatory fin locomotion. The Cuttlebot has four primary artificial muscles in its fins in addition to a tentacle-inspired soft gripper. The robot was evaluated in a series of tethered and untethered swimming tests, demonstrating a top speed of 2.6 centimeters per second translation and 10 degrees per second rotation. Furthermore, the CORE system was capable of driving specialized control signals into the artificial muscles to controllably output force and torque in six axes. This work provides a platform for developing complex, bio-inspired swimming robots for ocean exploration and monitoring, laying the foundation with our leading example: the Cuttlebot.

**One-Sentence Summary:** An autonomous, reconfigurable robotic platform enables the development of a complex cuttlefish-inspired swimming robot.

**Main Text:**

## INTRODUCTION

Living, moving systems can be found across all extremes of the ocean, from hydrothermal vents to icebergs, from the sunlit surface to the bottom of the Mariana trench, but underwater robots have limited reach because of their actuators. Increased interest in deep-sea and other extreme environment operations motivates the development of a diverse range of robots to perform tasks that humans cannot, and replace humans in environments that would put us at risk (1, 2). Soft robots are very well suited to operate in unstructured environments and interact with sensitive systems and ecosystems. In nature, such capabilities are fundamental and incredibly varied; in comparison, most robotic demonstrations to date are flapping fins (3, 4, 5, 6). Typical soft robotics research develops from the invention or analysis of a soft mechanism, around which a supporting robotic infrastructure is built. Developing soft robotic appendages out of a central system is only possible with versatile artificial muscles.

To operate across the great span of ocean temperatures and pressures, a robot's actuators must be pressure and temperature-agnostic. Among varied classes of soft actuators, dielectric elastomer actuators (DEAs) stand out for deep sea applications due to their pressure independence, high deformation, energy density, and bandwidth (7). DEA-driven robots have been deployed from the furthest depths of the ocean (8) to the stratosphere (9). Consider the composition of natural muscle systems. The microstructure of animal muscle – specifically, muscular hydrostats – consists of a three-dimensional matrix of muscle fibers and connective tissues (10), with individual fibers approximately 5-100 μm in diameter and varying lengths up to a few centimeters. Muscular hydrostats are fully soft, but can rapidly change their morphology and stiffness through select and antagonistic control within the muscular matrix (10). An interesting example of this structure can be found in cuttlefish fins, which produce undulatory waves for precise swimming in three dimensions (11). Comparatively, DEAs consist of a dielectric elastomer, usually of 10-100 μm thickness, which separates two compliant electrodes (12). Under driving voltage, the DEA will rapidly contract in the axis perpendicular to the electrodes, expanding radially in the electrode plane. Established manufacturing techniques allow three-dimensional, multilayered structures to be fabricated (7), and strain limiting can be integrated akin to connective tissue to promote specific axes or morphologies of deformation. Natural muscle has the advantage in structural complexity and internal, on-demand chemical energy that can be converted to mechanical energy. However, DEAs win out in extreme temperature performance (9) and higher specific energy (13, 14), dependent on material selection. Utilizing the structural similarities to natural muscle and advantageous fundamental properties, we developed a cuttlefish-inspired, DEA-driven fin capable of 3D swimming through the depths of the ocean.

DEAs require driving voltages in the kilo-volt range, limiting robotic demonstrations to low degrees of freedom or tethered designs. The system integration challenge arises from the need for a unified power and control structure which can support multiple actuators in parallel under strict space and weight requirements. Conventional hydraulic and pneumatic actuators face significant challenges to deployment in the deep sea, primarily from their bulky driving hardware and the challenge of generating a pressure differential within high pressure environments (2). Other technologies like SMAs are too slow or difficult to reverse (15), and LCEs are highly environmentally sensitive and slow to actuate (16, 17). Electrically driven actuators minimize the amount of requisite supporting hardware and simplify electromechanical system modelling. Still, underwater DEA robots have been hindered by the need for multiple high voltage converters (5, 18), pre-stretched actuators (8), and potentially leaky electrohydraulic actuator designs (18, 19).

This work introduces CORE: a Centralized Operations Robot Envelope. CORE is an autonomous platform with integrated sensing, power, processing, and modular high voltage control lines capable of driving DEAs of varying scale and nearly unlimited configurations. The system is primed to drive multifunctional subsystems from broad bandwidth high voltage lines, enabling a single line to drive multiple systems or mechanisms operating in distinct bandwidths. To best apply the CORE, we looked to nature for an organism which exhibits intelligence, versatility, and soft mechanism locomotion and manipulation. On this platform, we developed the Cuttlebot – a cuttlefish-inspired autonomous underwater robot. The Cuttlebot incorporates a traveling wave fin subsystem capable of 3D swimming and a tentacle-inspired soft gripper. The Cuttlebot swims at 0.1 bl/s tethered and 0.05 bl/s untethered with a turning speed of 10.17 deg/s, and is demonstrated identifying and tracking different objects, gripping them, and moving them in a tank. A diving demonstration shows the Cuttlebot's capacity for swimming and maneuvering in three dimensions using the fin subsystem.

The novel DEA-based subsystems on the Cuttlebot warrant deeper analysis and demonstrate the CORE platform's value in facilitating research on novel robotic mechanisms and systems. Several swimming modes were evaluated with this method including flow and no-flow scenarios. The CORE system allows soft mechanisms to be seamlessly integrated and swapped out, serving as a strong platform for further research and development of soft aquatic robotic systems.

## RESULTS

### The Cuttlebot

We present an untethered, autonomous, DEA-driven swimming robot: the Cuttlebot (Fig. 1H). Our aim was to design a robot driven by multiple independent soft systems with complementary performance, with future potential for multifunctional system integration and extreme environment operations. In the design and integration of soft systems, nature is our guidepost. Common cuttlefish (*sepia officinalis*) are highly intelligent, versatile creatures with features which have inspired broad-ranging research, particularly in the field of DEAs; they swim using undulating wave pectoral fins (20); manipulate objects with their tentacles (21); communicate using their skin (22); facilitate jet propulsion with a siphon (23); and control buoyancy with chemical diffusion in the cuttlebone (24). Prior research has proposed cuttlefish-inspired robots with fixed fin morphologies using rigid mechanisms (25, 26). The Cuttlebot integrates two core functionalities: three-dimensional swimming with undulating wave pectoral fins (Figs. 1A and 1B), and object manipulation with a soft gripper system (Figs. 1C and 1D). The CORE platform facilitates vision sensing and IMU data collection, and controls five independent high voltage lines (four lines for the fin system, one line for the gripper), enabling the Cuttlebot to track objects, swim after them, and move them following onboard programming.

The undulating wave pectoral fins consist of of two DEA bending actuators spanned by a passive flexible membrane, showcased in Fig. 1B. The mechanism of bending is shown in Fig. 1G. There are two fin modules, one on either side of the Cuttlebot. Excitation of a single bender by a periodic voltage propagates a wave through the passive membrane (Fig. 1F). Through different actuator combinations, periodic excitement allows the Cuttlebot to propel itself in x and y, and rotate in the xy plane. Asymmetric excitation (such as ramp signal) of the fin actuators unlocks three new axis of swimming – vertical swimming in z, and rotation in xz and yz planes. The Cuttlebot is designed with slightly biased internal mass to maintain upright posture underwater. The gripper system consists of four inversely-actuated benders connected in parallel. The natural state of the gripper is closed, and a voltage is applied to open it.

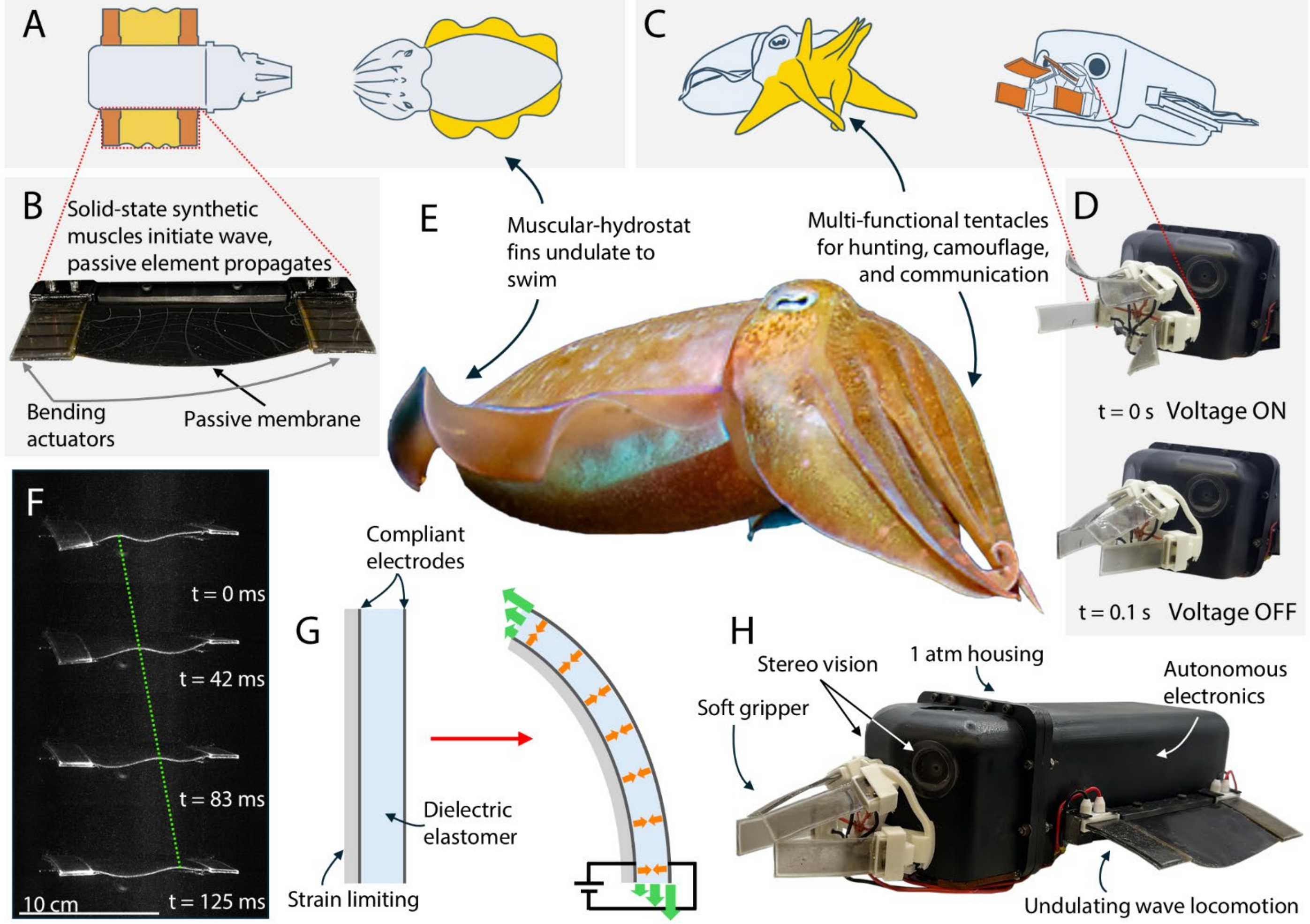

**Fig. 1. Untethered, cuttlefish-inspired robot design and functionalities.** (**A**) The Cuttlebot opposite a common cuttlefish (*sepia officinalis*), with mechanisms for undulatory wave fin swimming highlighted.(**B**) Diagram of a single undulatory wave swimming module with two actuators spanned by a passive element. (**C**) A cuttlefish spreads its tentacles opposite the Cuttlebot, opening its soft gripper. (**D**) The soft gripper mechanism is demonstrated in open (voltage on) and closed (voltage off) positions, with a closing time of 100 ms. (**E**) **Species** cuttlefish is shown, with muscular hydrostat structures highlighted. On the left are the undulating wave fins, which allow fin three dimensional maneuvering; on the right are the multi-functional tentacles. (**F**) The Cuttlebot fin module undulating wave is shown. Under periodic bending at 6 Hz, an actuator excites a traveling wave that propagates through the passive fin element. (**G**) Diagram of the working principal of a bending DEA actuator, of the same type used in the Cuttlebot. (**H**) The Cuttlebot assembly is shown, with integrated vision, autonomous electronics, and bio-inspired fin propulsion and soft gripper.

### *Robot design and fabrication*

The CORE Platform integrates sensing, processing, autonomous power, and multiple DOF actuator control. Two cameras provide 120° of vision, recognizing internally programmed objects and sending object data to the microprocessor. An integrated IMU provides positional feedback, allowing the CORE to implement closed-loop operations. Two custom PCBs house the electronics, with the lower board principally running low-voltage operations, and the upper board facilitating multi-line high-voltage control. The full electronic assembly is shown in Fig. 2A, and Fig. 2C shows an electrical schematic of a single HV control line. A full electrical schematic may be found in supplementary Fig. S1.

The Cuttlebot houses electronic components in a 3D-printed 1-atm housing of similar design methodology to Motsenbocker et. al (27). The housing was designed with a number of press-fit adapters, allowing DEA system modules to be rapidly reconfigured or replaced. The full assembly is shown in Fig. 2B. In pursuit of modularity we developed a method for electrically connecting and insulating bending actuators in compact, press-fit housings, shown in Fig. 2D. Thus, it was simple to replace entire fin modules when desired, or swap out a single actuator while keeping the rest of the sub-assembly intact.

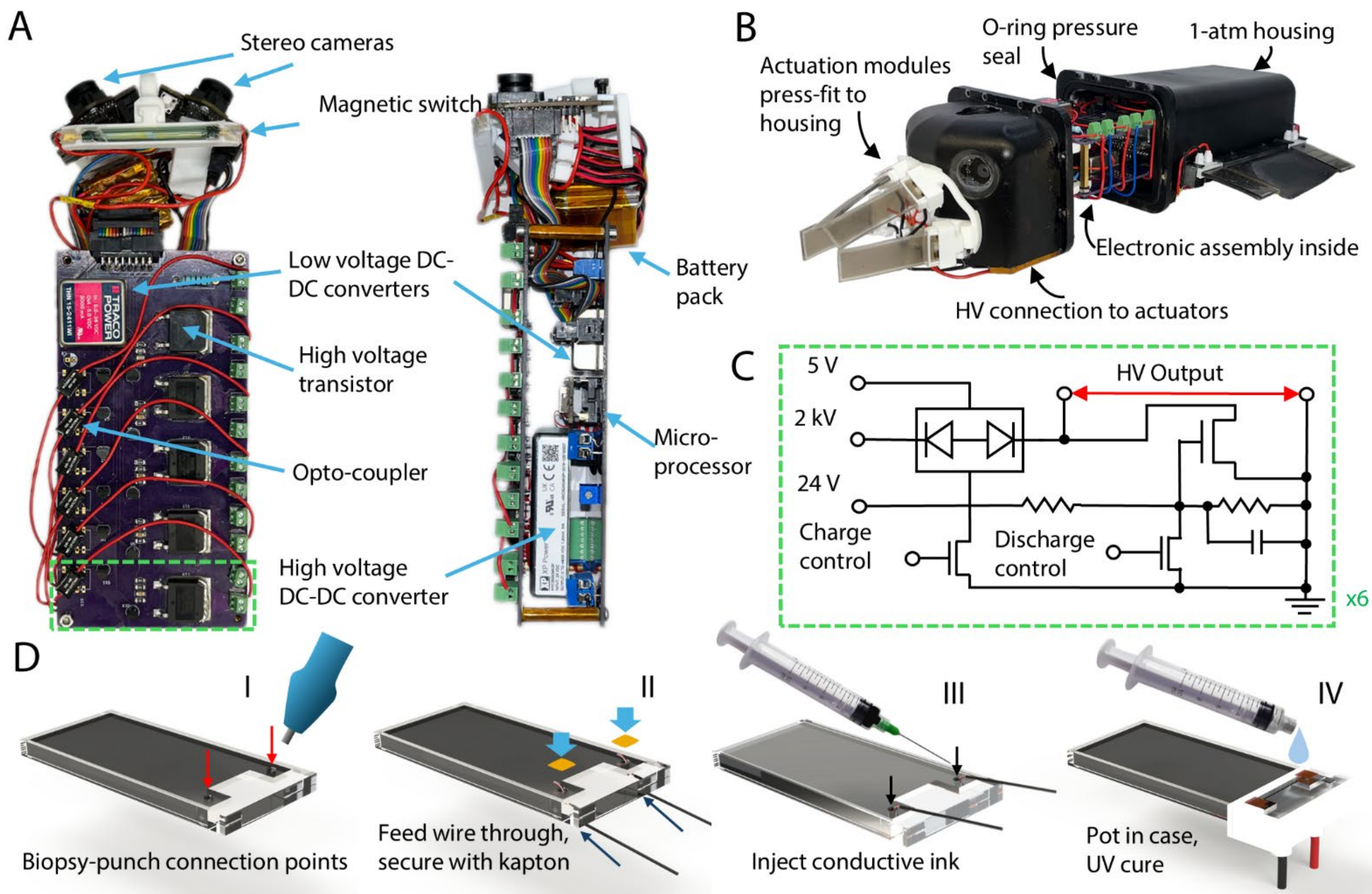

**Fig. 2. System design of the CORE platform and the cuttlebot.** (**A**) Two-level PCB layout of the autonomous CORE system. (**B**) High-level assembly showing autonomous electronics, one atmosphere housing, and actuator components. (**C**) Electrical schematic of the HV voltage control lines built into the CORE. (**D**) Actuator waterproofing diagram with all materials and assembly architecture.

### Three-dimensional swimming performance characterization

Tethered swimming performance was evaluated in near-surface swimming tests. The locomotion modules were fully submerged in all testing. DEA benders were made from an acrylic elastomer, CN9018 7.5% HDDA, with stamped CNT electrodes. The actuators measured 4 cm by 2 cm active area with a 2.8 mm thickness, and in all tests were driven by a periodic 2 kV square wave. The potting procedure integrating a DEA bending actuator into a waterproof, press-fit unit is shown in Fig. 2D. The passive fin element measured 4 cm by 9 cm with a 1 mm thickness. The Cuttlebot achieved a maximum forward (surge) swimming speed of 0.11 bl/s and a maximum transverse (sway) speed of 0.07 bl/s. Z-axis (yaw) rotation of over 10°/s was recorded (shown in supplementary video S3 and supplementary data S1).

We argue that the presented fin locomotion module is capable of 6-axis thrust vectoring through sequencing of periodic and asymmetric driving voltages. To validate this claim, the Cuttlebot was mounted to a six-axis force/torque (F/T) sensor, and subjected to excitation by rising and falling sawtooth waveforms in addition to the symmetric, periodic waveforms used in 2D swimming. The full array of excitation arrangements and resultant torques and forces are shown in Figs. 3C, 3E, and 3F, demonstrating that desired force/torque axes can achieve primary dominance through HV waveform control. Roll, pitch, and heave locomotion are driven by ramped electric fields corresponding to a gradual bending (or straightening) of the actuator, ending with a rapid switch in electric field magnitude and bend angle. Figures 3E and 3F show that this driving scheme yields torque and force measurements with high initial magnitude in the target vector followed by a longer period of minor oscillations with diminishing magnitude (not unlike a damped system's impulse response). Yaw, surge, and sway are driven by periodic square wave electric field inputs. The measured primary surge force shows regular periodic oscillation. However, the sway-swimming dominant force shows a higher peak at the start of the cycle (when the electric field is applied) and a second, lower peak when the field is removed. This could be due to lower force output during the actuator's elastic restorative actuation compared to active driving, or due to mechanical interactions between the actuators and the passive fin. The most surprising result was Fig. 3E for yaw-control, in which the yaw torque displayed lower peak magnitudes compared to the other two axes. This was particularly surprising because of the high rate of turning (over 10°/s) achieved during tethered testing (shown in supplementary video S3 and supplementary data S1). Off-axis forces and torques are likely due to discrepancies in actuator manufacture and module assembly, or unanticipated hydrodynamic interactions between the fin modules and Cuttlebot body. All measurements can be found in supplementary data S1 and S2.

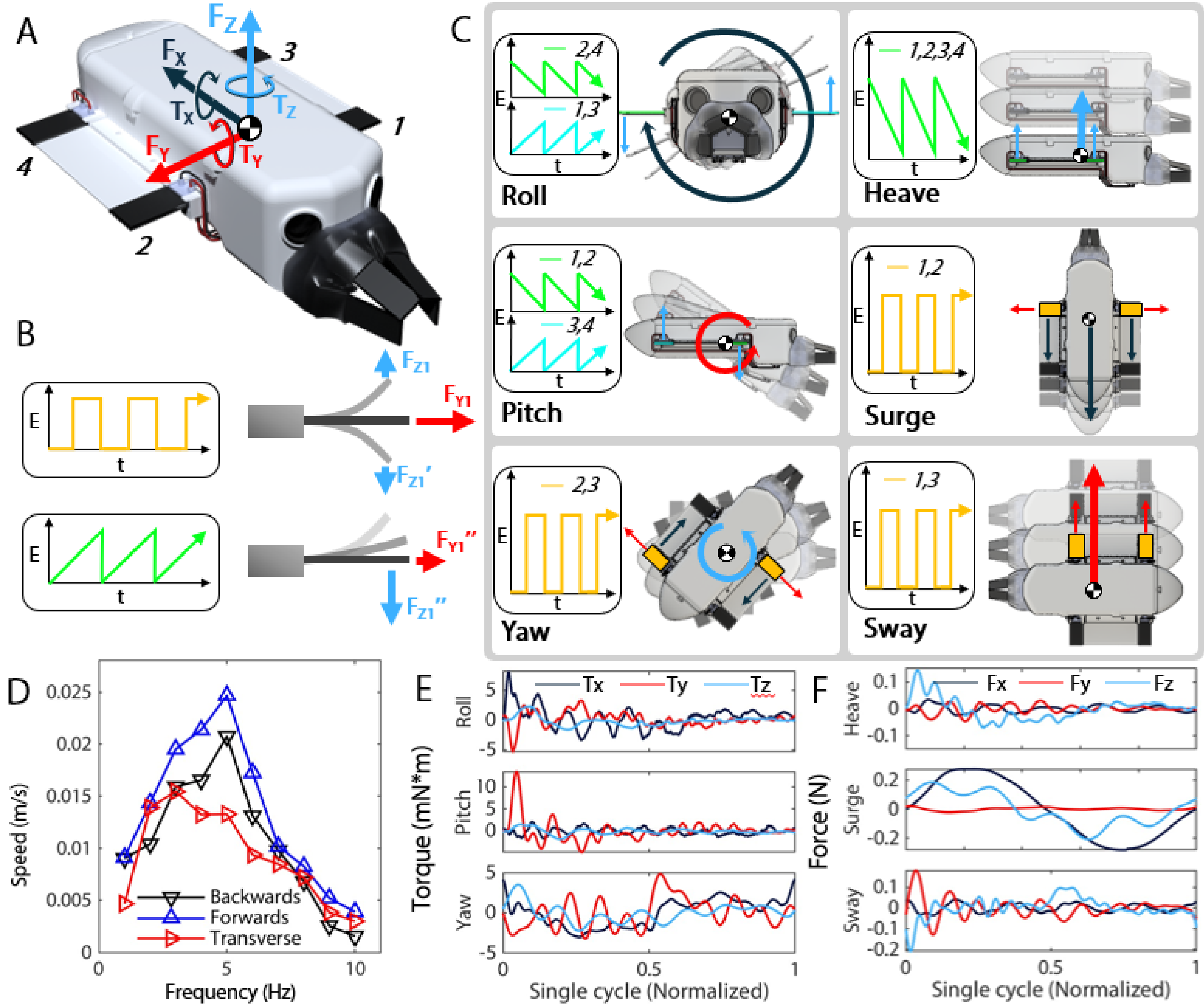

**Fig. 3. Different actuator sequences and waveforms enable 3D swimming.** (**A**) The fin is shown with force vectors corresponding to desired and measured quantities. Actuators are numbered 1-4, which are referenced in sub-plot legends in (C). (**B**) Periodic square and sawtooth HV signals result in production of different predominant force vectors. Subplots show applied electric field E over time t. (**C**) Diagram of all actuator sequence and control methods to achieve 6-axis force/torque control. (**D**) Tethered free-swimming speeds of a tethered Cuttlebot show highest forward and backwards speed at 5 Hz and highest transverse speed at 3 Hz. All actuators were driven at 2 kV under periodic square waves. Error bars determined by standard deviation of 10 s interval around peak recorded speed. (**E**) Torque measurements under periodic actuation as indicated in C. All modules run at 2 kV, roll and pitch at 1 Hz, and yaw at 2 Hz. Plots represent average of all tests under steady-state performance, n=10 for roll and pitch, n=20 for yaw. (**F**) Force measurements under periodic actuation as indicated in C. All modules run at 2 kV, surge at 4 Hz, heave and sway at 1 Hz. Plots represent average waveform of all tests under steady-state performance, n=10 for heave and sway, n=40 for surge.

### Autonomous operation

The autonomous, untethered Cuttlebot running off CORE was programmed to swim while tracking and following a target, actuating the soft gripper subsystem once close enough to grasp the target before retreating. This would be an analog to conducting an object retrieval for a user in the field, or identifying and capturing a biological sample for further examination (shown in Fig. 4A). Additionally, untethered testing validated the Cuttlebot's capacity for 3D swimming by driving synchronous, ramped voltages through all of the fin actuators in the manner described in Fig. 3C Heave. An alternating sequence of 10 "down" ramp actuations followed by 10 "up" ramp actuations were shown to respectively increase and decrease the robot's depth in the tank, shown in Fig. 4B and supplementary video S4.

Towards future deployment in extreme environments, a set of multilayer actuators were subjected to submerged pressurization in a pressure chamber up to 3000 psi, corresponding to an ocean depth of approximately 2000 meters.The actuators showed no decrease in performance following exposure (bending performance before and after pressurization shown in supplementary Fig. S2, supplementary data S3). Autonomous swimming was conducted in the field at two locations representing potential future applications. The Cuttlebot could be used as a biologically-sensitive platform for researching coastal ecosystems, shown in Fig. 4C swimming off the coast of Honolulu, HI. Another use case would be in submerged, radiation-exposed environments, where the inherent pressure tolerance and broad temperature band of DEAs make them potentially suitable for future nuclear power operations. Figure 4D shows the Cuttlebot swimming in a spent nuclear fuel pool, showcasing potential as an inspection system for difficult-to-access irradiated environments.

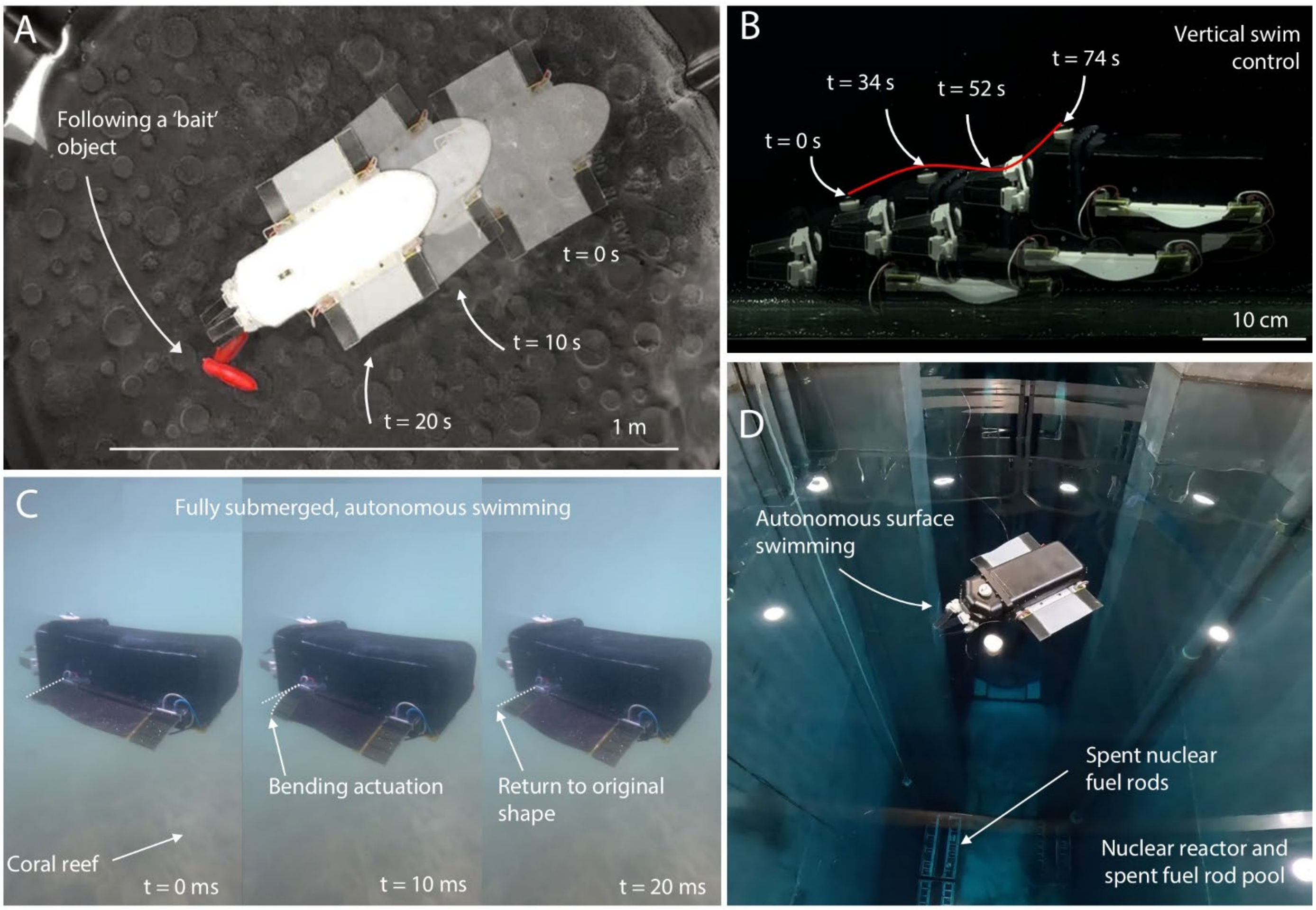

**Fig. 4. Autonomous robot operation showcasing integrated capabilities and different environments.** (**A**) The Cuttlebot swims in 2D space, tracking a target. Once near, it grasps the target, then reverses away. (**B**) 3D swimming is enabled through actuator control as described in Fig. 3C *Heave*. The autonomous robot performs 20 s of vertical swimming, 20 s of downward swimming, and 20 s upward swimming with 5 s break between modes. (**C**) The Cuttlebot swims autonomously in a coral reef off the coast of Waikiki, HI. (**D**) The Cuttlebot swims autonomously in a spent nuclear fuel rod pool, highlighting potential future deployment in radioactive, pressurized environments.

## DISCUSSION

We presented CORE: a Centralized Operations Robot Envelope, and the Cuttlebot, a cuttlefish-inspired robot developed on the CORE platform. The robot was capable of integrating power autonomy, sensing, and high-DOF actuator control to respond to its environment, swim in three dimensions, and carry out programmed tasks. The fin subsystem designed for the Cuttlebot was evaluated in tethered swimming speed tests and blocked F/T measurements, and autonomous demonstrations showcased the robot's capacity for integrating soft actuator functionalities into the modular, high-DOF platform.

Through a series of surface swimming experiments, the Cuttlebot showed the ability to navigate within a 2D plane with highly controllable directional swimming and rotation, with a top speed of 0.11 bl/s and xy-plane rotation over 10°/s. The robot was also demonstrated controlled, untethered depth control. Force and torque measurements of the robot under various swimming modes demonstrated that the suggested driving modes can isolate desired force/torque vectors. Hydrodynamic analysis of the fin mechanism could provide insights to the specific fluid-fin interactions governing thrust and torque under the swimming modes. This analysis could be achieved with synchronized F/T and PIV measurements about the robot during various swim modes, similar to (28). Preliminary Stereo-PIV was performed on a fin module during forward and transverse locomotion modes. Further information is available in supplementary materials and methods and supplementary figure S3.

Full utility of the undulating fin propulsion mechanism is locked behind precise ballasting of the robot. With slightly negative or positive buoyancy, the Cuttlebot is able to adjust its depth as well as navigate in 2 dimensions within its plane, enabling effective 3D swimming. However, the Cuttlebot is unable to meaningfully adjust its roll or pitch within the water due to its non-uniform density. Within a highly compact 1-atm housing, with components of set mass, density, and layout, as well as human variability in the manufacture and assembly process, fabrication of a finely balanced and ballasted swimmer is a delicate challenge.

High voltage connections in underwater robotics remain a challenge, particularly in systems where commercial connectors' bulk would severely impact the robot's performance. Some leakage current into the surrounding fluid causes inefficiencies and potential failures if not mitigated. The presented work took steps towards development of pressure-resilient potted DEAs, but the matter of how to best connect devices to internal electronics remains an open problem.

DEAs, despite requiring complex fabrication, offer high potential for multifunctionality via material selection and inherent properties. For example, integration of electroluminescent powders into the dielectric layer enables integrated, multi-band lighting or communication into the artificial muscle (29). DEAs also may be used as capacitive energy storage for load balancing, and that same capacitance contains information about their morphology for self-sensing (30). Small DEAs have been used as multifunctional actuators and speakers, capable of producing both effects independently and concurrently (31). Many of these capabilities inspire comparison with the Cuttlebot's inspiration – the cuttlefish. Cephalopod-inspired robotic siphons have been investigated (32, 33), and could introduce a multimodal locomotion scheme with quick, agile locomotion facilitated by the siphon and efficient, precise control provided by undulating wave fins. Full utilization of these extended and inherent capabilities has the potential to extend the autonomy and functionality of future DEA-powered robots built on the CORE platform.

Use of a microprocessor severely limits the processing, complexity, and autonomy of systems like Cuttlebot built on the CORE. Trivial adjustments to the CORE platform could replace the microprocessor with a single board computer (e.g. raspberry pi, Jetson Nano), enabling highly complex programming and AI implementation like computer vision to the system. This would significantly extend the autonomy, field utility, and research capacity of future systems built on the CORE.

## MATERIALS AND METHODS

### Dielectric elastomer actuator benders

The DEA benders used in this study were based on an established multilayering process presented in Duduta *et al* (7). The elastomer was CN9018 (Sartomer), and the electrodes were fabricated in the method described in Duduta *et al* to an CNT ink areal density of 8.13 $\mu L/cm^2$. Devices were strain limited with a 12.5 µm mylar layer, and uni-axial bending was promoted by integrating carbon-fiber rods parallel to the bending axis into the strain limiter. Electrical connection points were made by hole-punch and infilled with a conductive fluid medium to interface with the wire connections. The devices were then integrated into 3D printed housings and potted with the same CN9018 elastomer. Following this processing method, each device was a modular, plug-in unit capable of repeatable, long-term underwater actuation. This methodology also allowed for rapid swapping and prototyping of locomotion unit designs.

### Design of traveling-wave locomotion module

The cuttlefish-inspired fin locomotion system is comprised of two symmetric modules, each of which is comprised of a 3D-printed platform, a modular passive fin element, and two DEA bender modules affixed to the ends of the passive element. The fin modules may each be press-fit to the main body of the Cuttlebot, allowing for whole modules or individual actuators to be easily replaced and inter-changed. Further description of the module assembly may be found in the supplementary materials and methods.

### Design of waterproof housing

The CORE platform is contained in a 3D printed 1 atm housing (Formlabs 4L printer, clear resin v5). Two shell components are bolted together with a buna-n o-ring providing a seal. High voltage through-shell electrical connections to provide power to the DEAs were potted with TotalBoat Thickset epoxy and left to cure for 4 days before submersion. Further information is included in the supplementary materials and methods.

### Design of CORE platform

CORE was designed as a unified platform integrating sensing, high voltage control, and autonomous power. It is built entirely from off-the-shelf components soldered to a custom

PCB. CORE uses two low-power cameras with in-built object recognition (Pixy2), providing 120° field-of-view combined. An IMU integrated into the microprocessor (Arduino Nano RP2040 Connect) provides the robot with spatial awareness. A 5 W high voltage DC-DC converter supplies a controllable voltage of 0 - 4 kV (XP Power HRC0524S4K0P) to six high-voltage switching circuits, each capable of driving independent DEAs or interconnected DEA systems (such as used in the gripper). The high voltage switching circuits operate on two inputs which control an opto-coupler (OR-100) for charging and a HV MOSFET (IXTT02N450HV) for discharging the line. Figure 4C shows the full charge-discharge circuit, which is capable of producing analog output voltages between 0 V and the HV line voltage, and digital switching at over 1 kHz. The microprocessor integrates object data from the cameras, IMU data, and HV line outputs in control loops to accomplish the objectives demonstrated above. A six-cell Li-ion battery pack powers the CORE for up to an hour of autonomous operation, and low-voltage DC-DC converters (THN 15-2411WI) distribute power to circuit components at requisite levels. A full PCB schematic is included in supplementary Fig. S1.

## Characterization of Cuttlebot swimming

### *Tethered 2D swimming characterization*

The tethered robot was operated in a circular tank of 1.2 meters diameter, filled to a depth of 18 cm. The actuators were driven by a high voltage power supply (Trek 610E High Voltage Amplifier) triggered by a signal generator (insert), driving the actuators with square-wave waveforms. Kinovea software was used to evaluate swimming speed and robot trajectories.

### *Vertical swimming capability*

The untethered Cuttlebot was submerged in a 120 cm by 45 cm tank with a depth of 60cm, and trimmed to marginally negative buoyancy. An "upward swim" and a "downward swim" protocol were cyclically driven, each consisting of ten asymmetric fin actuation cycles (Fig. **XB**) at 0.5 Hz followed by a five second rest. Path tracking swim speed measurements were evaluated using Kinovea software.

### *Blocked force/torque measurements*

The tethered robot was mounted to a six-axis force/torque sensor (ATI mini40) and submerged to a depth of 10 cm (measured from the surface to the top of the fin module) in a recirculating water channel with 45cm width, 2.4 m length, and 30cm depth. Force/torque measurements were collected via DAQ at a rate of 1 kHz, and smoothed using outlier removal and moving average filter. In tests where the water channel flow was on, the fluid velocity was set to 4 cm/s to approximate the robot's forward swim speed. Actuators were driven by identical equipment to the tethered 2D swimming tests, but additional, non-symmetric waveforms were generated to enable force/torque prioritization for three axes not yet characterized; vertical force, roll torque, and pitch torque. Roll and pitch tests were driven by the CORE platform, as they required multiple complex high voltage outputs. Otherwise, testing was driven by a desktop high voltage amplifier and signal generator (Trek 610E).

## Statistical Methods

Tethered swimming speed measurements were performed on a single robot prototype filmed at 30 frames per second, and speed was calculated in *Kinovea* software and exported to Excel. A 30-frame simple moving average (SMA) filter was applied to the data prior to selecting a peak value, and error bars were calculated as standard deviation of n=50 frames of unfiltered speed measurements about the peak frame. Force/torque measurements were sampled at 1 kHz and processed in MATLAB with outlier removal, 100-frame SMA filter, and presented as the point-by-point average of all cycles recorded in a given test (roll $f$=1 Hz, n=10; pitch $f$=1 Hz, n=10; yaw $f$=2 Hz, n=20; heave $f$=1 Hz, n=10; surge $f$=4 Hz, n=40; sway $f$=1 Hz, n=10).

## Supplementary Materials

List the titles of the Supplementary Materials in the following order: Supplementary materials and methods (if necessary), supplementary figures, supplementary tables, then other supplementary files such as movies, data, interactive images, or database files. Be sure to submit all Supplementary Materials with the manuscript. **Re-order in accordance with reference appearance in manuscript.**

Materials and Methods

- Preliminary stereo-PIV on the fin subsystem
- Fabrication of 1-atm housing
- Fabrication of fin locomotion module

Fig. S1 or Figs. S1 to Sx for multiple supplementary figures

- Fig. S1: CORE board schematic
- Fig. S2: Actuator bending performance before and after high-pressure submersion
- Fig. S3: Stereo-PIV of the robotic undulating wave fin
-

Movie S1 or Movies S1 to Sx for multiple supplementary movies

- Movie S1: Tethered forward swimming
- Movie S2: Tethered transverse swimming
- Movie S3: Tethered yaw-control swimming
- Movie S4: Untethered vertical swimming

Data S1, S2

- Data S1: Swim speed measurements
- Data S2: Blocked force/torque measurements
- Data S3: Performance of bending actuators before and after applied pressure

## References and Notes

1. D. O. B. Jones, M. B. Arias, L. V. Audenhaege, S. Blackbird, C. Boolukos, G. Bribiesca-Contreras; J. T. Copley, A. Dale, S. Evans, B. F. M. Fleming, A. R. Gates, H. Grant, M. G. J. Hartl, V. A. I. Huvenne, R. M. Jeffreys, P. Josso, L. D. King, E. Simon-Lledó, T. L. Bas, L. Norman, B. O’Malley, T. Peacock, T. Shimmield, E.V. D. Stewart, A. K. Sweetman, C. Wardell, D. Aleynik, A. G. Glover, Long-term impact and biological

recovery in a deep-sea mining track. *nature*. 642, 112-118 (2025). doi: 10.1038/s41586-025-08921-3

2. G. Li, T. Wong, B. Shih, C. Guo, L. Wang, J. Liu, T. Wang, X. Liu, J. Yan, B. Wu, F. Yu, Y. Chen, Y. Liang, Y. Xue, C. Wang, S. He, L. Wen, M. T. Tolley, A. Zhang, C. Laschi, T. Li, Bioinspired soft robots for deep-sea exploration. *nat. commun.* 14, 7097 (2023). doi: 10.1038/s41467-023-42882-3
3. R. K. Katzschmann, A. D. Marchese, D. Rus, Hydraulic Autonomous Soft Robotic Fish for 3D Swimming, in *Experimental Robotics*, M. Hsieh, O. Khatib, V. Kumar, Eds. (Springer International Publishing, Cham, 2015).
4. S. Wang, B. Huang, D. McCoul, M. Li, L. Mu, J. Zhao, A soft breaststroke-inspired swimming robot actuated by dielectric elastomers. *Smart Mater. Struct.* 28, 045006 (2019). doi: 10.1088/1361-665X/ab0a7a
5. F. Berlinger, M. Duduta, H. Gloria, D. Clarke, R. Nagpal, R. Wood, A Modular Dielectric Elastomer Actuator to Drive Miniature Autonomous Underwater Vehicles, in Proceedings of the 2018 IEEE International Conference on Robotics and Automation, 21 to 25 May 2018, Brisbane, QLD, Australia, pp. 3429-3435.
6. R. Wang, C. Zhang, Y. Zhang, W. Tan, W. Chen, L. Liu, Soft Underwater Swimming Robots Based on Artificial Muscle. *Adv. Mater. Technol.* 8, 2200962 (2023). doi: 10.1002/admt.202200962
7. M. Duduta, E. Hajiesmaili, H. Zhao, R. J. Wood, D. R. Clarke, Realizing the potential of dielectric elastomer artificial muscles. *Proc. Natl. Acad. Sci.* 116, 2476-2481 (2019). doi: 10.1073/pnas.1815053116
8. G. Li, X. Chen, F. Zhou, Y. Liang, Y. Xiao, X. Cao, Z. Zhang, M. Zhang, B. Wu, S. Yin, Y. Xu, H. Fan, Z. Chen, W. Song, W. Yang, B. Pan, J. Hou, W. Zhou, S. He, X. Yang, G. Mao, Z. Jia, H. Zhou, T. Li, S. Qu, Z. Xu, Z. Huang, Y. Luo, T. Xie, J. Gu, S. Zhu, W. Yang, Self-powered soft robot in the Mariana Trench. *nature.* 591, 66-71 (2021). doi: 10.1038/s41586-020-03153-z
9. C. Tugui, T. Thakar, A. Gogoj, A. White, A. Li, A. Yin, E. Pomianek, M. Duduta, A Soft Robotic Demonstration in the Stratosphere. arXiv:2603.04352 [cs.RO] (2026).
10. W. Kier, The diversity of hydrostatic skeletons. *J. Exp. Biol.* 215, 1247-1257 (2012). doi: 10.1242/jeb.056549
11. W. Kier, The fin musculature of cuttlefish and squid (Mollusca, Cephalopoda): morphology and mechanics. *J. Zool., Lond.* 217, 23-38 (1989). doi: 10.1111/j.1469-7998.1989.tb02472.x
12. R. Pelrine, R. Kornbluh, Q. Pei, J. Joseph, High-Speed Electrically Actuated Elastomers with Strain Greater Than 100%. *Science*. 287, 836-839 (2000). doi: 10.1126/science.287.5454.83
13. W. Feng, L. Sun, Z. Jin, L. Chen, Y. Liu, H. Xu, C. Wang, A large-strain and ultrahigh energy density dielectric elastomer for fast moving soft robot. *Nature Communications.* 15, 4222 (2024). doi: 10.1038/s41467-024-48243-y
14. Y. Shi, E. Askounis, R. Plamthottam, T. Libby, Z. Peng, K. Youssef, J. Pu, R. Pelrine, Q. Pei, A processable, high-performance dielectric elastomer and multilayering process. *Science*. 377, 228-232 (2022). doi: 10.1126/science.abn0099
15. W. Chu, K. Lee, S. Song, M. Han, J. Lee, H. Kim, M. Kim, Y. Park, K. Cho, S. Ahn, Review of biomimetic underwater robots using smart actuators. *Int. J. Precis. Eng. Manuf.* 24, 1281-1292 (2012). doi: 10.1007/s12541-012-0171-7

16. J. Speregen, T. White, Liquid Crystalline Elastomers in Soft Robotics: Assessing Promise and Limitations. *Advanced Robotics Research*. Early View (2025) doi: 10.1002/adrr.202500150
17. T. Zang, J. Wang, G. Yan, X. Lu, J. Hu, H. Xia, Y. Zhao, State of the Art, Insights and Perspectives for Bio-Inspired Liquid Crystal Elastomer Soft Actuators. *Advanced Materials*. 37, e08694 (2025). doi: 10.1002/adma.202508694
18. F. Hartmann, M. Baskaran, G. Raynaud, M. Benbedda, K. Mulleners, H. Shea, Highly agile flat swimming robot. *Sci. Robot.* 10 eadr0721 (2025). doi: 10.1126/scirobotics.adr0721
19. G. Li, P. Shen, T. Wong, M. Liu, Z. Sun, X. Liu, Y. Chen, X. Wang, H. Zhang, B. Hu, D. Chen, Z. Zhang, C. Zhang, R. Wang, W. Zhang, S. Nie, X. Zhang, J. Wong, H. Zhou, W. Li, H. Wang, Q. Zhang, S. Wang, Z. Yu, H. Li, H. Zhao, Q. Zeng, S. Wang, Z. Huang, C. Ye, A. Zhang, T. Li, Plasticized electrohydraulic robot autopilots in the deep sea. *Sci. Robot.* 10 eadt8054 (2025). doi: 10.1126/scirobotics.adt8054
20. W. Kier, The fin musculature of cuttlefish and squid (Mollusca, Cephalopoda): Morphology and mechanics. *Journal of Zoology*. 217, 23-38 (1989). doi: 10.1111/j.1469-7998.1989.tb02472.x
21. W. Kier, The Musculature of Coleoid Cephalopod Arms and Tentacles. *Front. Cell Dev. Biol.* 4:10 (2016). doi: 10.3389/fcell.2016.00010
22. E. N. Shook, G. T. Barlow, D. Garcia-Rosales, C. J. Gibbons, T. G. Montague, Dynamic skin behaviors in cephalopods. *Current Opinion in Neurobiology*. 86:102876 (2024). doi: 10.1016/j.conb.2024.102876
23. G. J. Gemmell, J. O. Dabiri, S. P. Colin, J. H. Costello, J. P. Townsend, K. R. Sutherland, Cool your jets: biological jet propulsion in marine invertebrates. *J Exp Biol.* 224(12):jeb222083 (2021). doi: 10.1242/jeb.222083
24. E. J. Denton, 6 - The Buoyancy of Fish and Cephalopods. *Progress in Biophysics and Biophysical Chemistry.* 11, 177-234 (1961). doi: 10.1016/S0096-4174(18)30212-9
25. E. Arslan, K. Akça, A Design Methodology for Cuttlefish Shaped Amphibious Robot. *European Journal of Science and Technology.* Special Issue, 214-224 (2019). doi: 10.31590/ejosat.637838
26. Y. Wang, Z. Wang, J. Li, Initial Design of a Biomimetic Cuttlefish Robot Actuated by SMA Wires. *2011 Third International Conference on Measuring Technology and Mechatronics Automation*, Shanghai, China (2011). doi: 10.1109/ICMTMA.2011.393
27. B. Motsenbocker, O. Mattyasovszky, S. Mayer, B. Phillips, A Design Guide for Manufacturing Deep-Sea Pressure Vessels with SLA 3D Printing, in *OCEANS 2025 Brest* (2025), Brest, France, pp. 1-7. doi: 10.1109/OCEANS58557.2025.11104294
28. H. Ko, V. Saro-Cortes, B. Mmari, D. Ni, A. Wissa, R. Nagpal, BlueGuppy: tunable kinematics enables maneuverability in a minimalist fish-like robot. *Bioinspiration and Biomimetics.* 20 056006 (2025). doi: 10.1088/1748-3190/adf2e9
29. B. Sun, Y. Ling, K. Wang, Q. Zhang, Y. Li, H. Wang, K. Li, C. Hou, All-in-One Luminescent Dielectric Elastomer Actuator Driven by a Single Electrical Stimulus with Large Out-of-Plant Actuation and Fast Response. *Small*. 19 e10346 (2026). doi: 10.1002/smll.202510346
30. I. A. Anderson, T. A. Gisby, T. G. McKay, B. M. O'Brien, E. P. Calius, Multi-functional dielectric elastomer artificial muscles for soft and smart machines. *Journal of Applied Physics.* 112, 041101 (2012). doi: 10.1063/1.4740023

31. S. Gratz-Kelly, G. Rizzello, M. Fontana, S. Seelecke, G. Moretti, A Multi-Mode, Multi-Frequency Dielectric Elastomer Actuator. *Adv. Funct. Mater*. 32, 2201889 (2022). doi: 10.1002/adfm.202201889
32. D. Flores, S. Sandhu, A. White, A. Yin, A. L. Li, S. Kang, Y. Wang, L. P. Chamorro, M. Duduta, RoboNautilus: a cephalopod-inspired soft robotic siphon for underwater propulsion. *npj robotics.* 3, 17 (2025). doi: 10.1038/s44182-025-00035-2
33. R. Zhang, Z. Shen, H. Zhong, J. Tan, Y. Hu, Z. Wang, A Cephalopod-Inspired Soft-Robotic Siphon for Thrust Vectoring and Flow Rate Regulation. *Soft Robotics.* 8, 416-431 (2021). doi: 10.1089/soro.2019.0152

**Acknowledgments:**
We wish to thank Dr. Brennan Phillips at the University of Rhode Island for providing us with lab access for high-pressure testing. We also thank Dr. Codrin Tugui of the Petru Poni Institute of Macromolecular Chemistry for broad-ranging guidance.

**Author contributions:**

Conceptualization: MD, ANW
Methodology: ANW
Investigation: ANW, ALL, AHY, VSC, DR, HW
Resources: MD, AW
Visualization: ANW, MD
Funding acquisition: MD
Project administration: MD, ANW
Supervision: MD
Writing – original draft: ANW
Writing – review & editing: ANW, ALL, AHY, AW, MD

**Competing interests:** Include any financial interests of the authors (including but not limited to financial holdings, professional affiliations, advisory positions, and board memberships) that could be perceived as being a conflict of interest. Also include any awarded or filed patents pertaining to the results presented in the paper. When authors have no competing interests, this should also be declared (e.g., "Authors declare that they have no competing interests.").

**Data and materials availability:** All data are available in the main text or the supplementary materials.

## Supplementary Materials and Methods:

### Preliminary stereo-PIV on the fin subsystem:

Stereo-PIV was conducted on a single fin subsystem through a test matrix consisting of flow and no-flow conditions, various swimming modes, and image plane positions cross-sectioning out from the fin, within the fin area and in the wake of the fin. The PIV setup is shown in Fig. S1, as well as some preliminary results. The setup incorporates two 9-Megapixel Photron Nova R5 high-speed cameras, with LaVision Scheimpflug Adaptors attached. Image pairs were recorded with a 3 ms delay at an overarching acquisition rate of 48 Hz (allowing 8 and 12-frame PIV image cycles at fin frequencies of 6 and 4 Hz respectively). Figs. S1C and S1D showcase preliminary PIV analysis of forward fin swimming under flow and no-flow conditions, under image processing schemes described in the figure caption.

### Fabrication of 1-atm housing:

The Cuttlebot housing is a one-atmosphere water-tight housing consisting of two primary components – the cap and the shell. Both parts were designed in SolidWorks and printed with a Form 4L SLA printer from Clear Resin V5 (Formlabs). The mate surfaces were sanded on a flat surface to ensure planar regularity with increasingly fine grit sandpaper (220 to 400-grit).

The cap requires further processing. 1/8" clear acrylic sheet was laser-cut and inset to the viewports with a commercial epoxy (Sil-Poxy, Smooth-On, Inc). High-voltage, low-profile electrical connections were fabricated by potting two 6-row male pin headers into the cap using Fathom Deep Pour Epoxy Resin (TotalBoat).

The cap and shell mate face-to-face, and the shell part incorporates a groove that seats a Buna-N O-ring (cut in-house) . The two parts are assembled with M2.5 bolts, providing relatively quick access to the electronic interior for rapid prototyping.

### Fabrication of fin locomotion module:

The Cuttlebot has two identical fin subsystems, one on either side of the body. These fins allow the robot to 3D-swim underwater. Each subsystem consists of two potted DEA benders, the module base, the passive fin element, and the fin element clip. The passive element is secured between the module base and fin element holder with M2.5 bolts, and attached to the benders with kapton tape. The full locomotion module is press-fit to the Cuttlebot shell component, and actuator electrical connections plug into the cap component's potted electrical connection points.

## Supplementary figures:

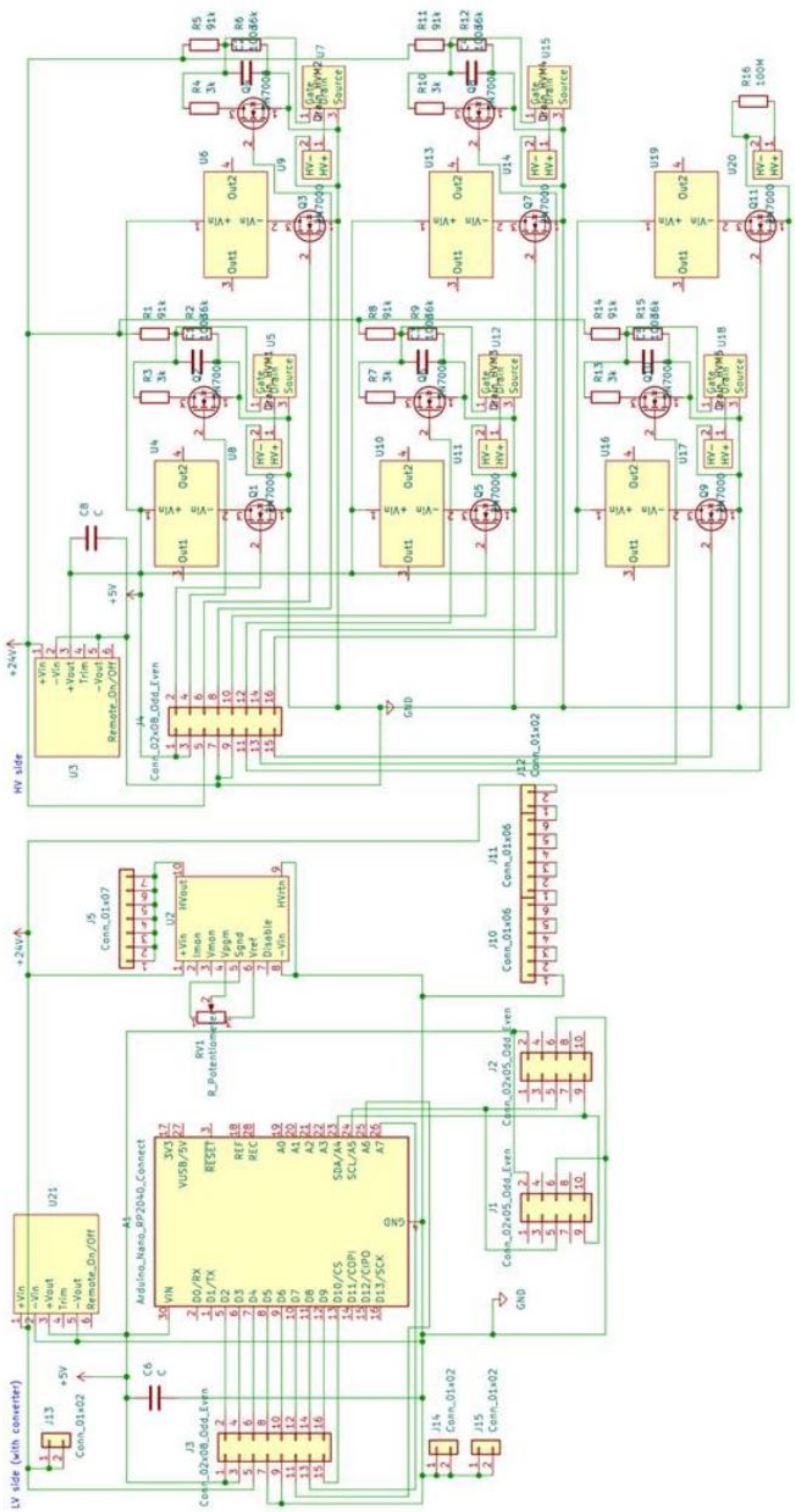

**Fig. S1. CORE board schematic.** An electrical schematic of the two-board CORE system, printed on two PCBs. J1, J2 route to Pixy v2 cameras (PixyCam). J3 and J4 connect via ribbon-cable to link the two PCB boards. U4, U6, U10, U13, U16, U19 represent opto-couplers, with HV

connections that attach to screw terminals J5 (HV in) and U8, U9, U11, U14, U17, and U20 (HV out).

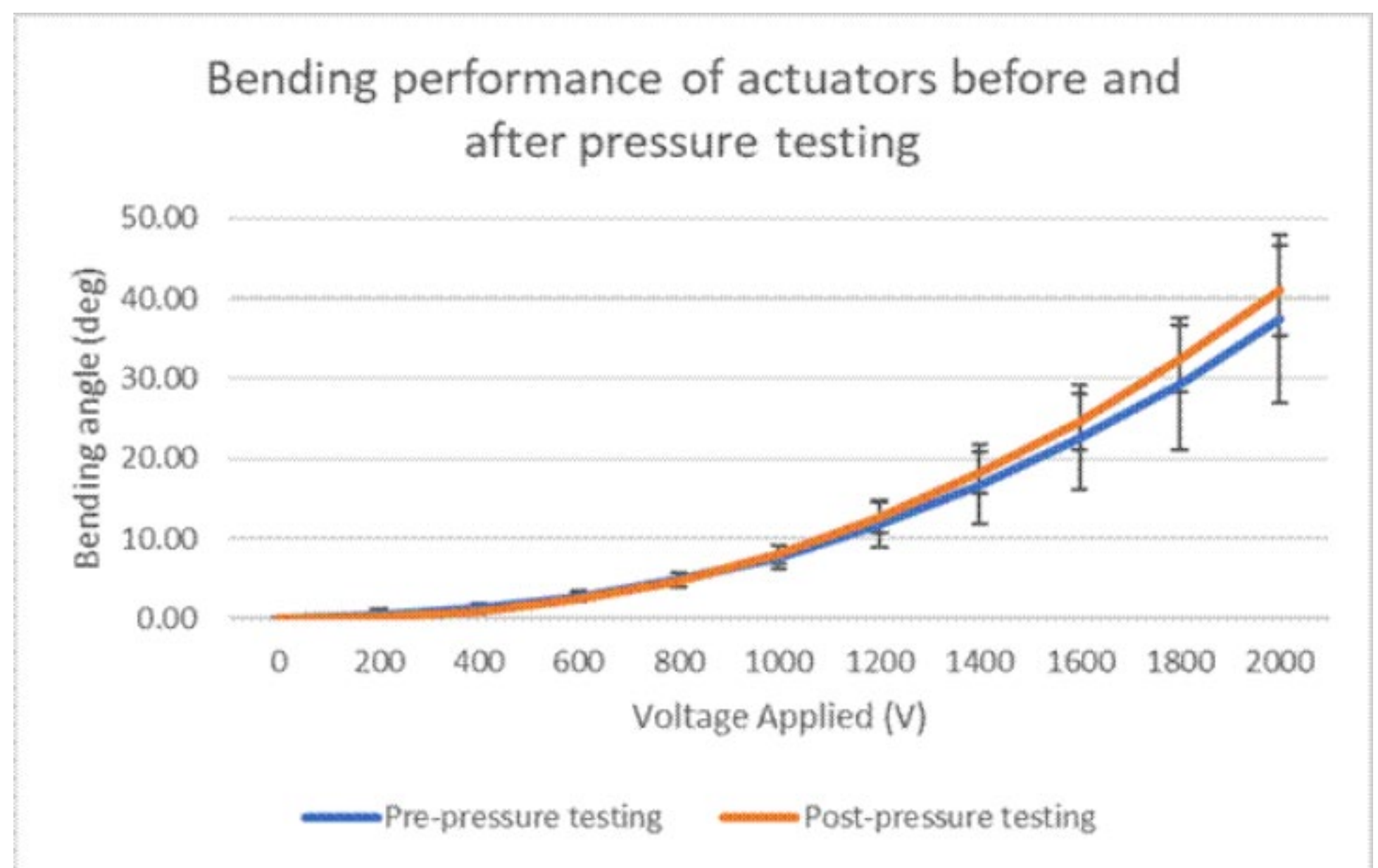

**Fig. S2. Actuator bending performance before and after high-pressure submersion.** A sample set of bending actuators (n=6) identical to those used in the Cuttlebot fins were subjected to applied voltages of 0 to 2 kV at intervals of 200 V. Bend angles were recorded with an overhead camera (DFK 37BUX250, IMAGINGSOURCE) with a zoom lens (ZOOM 7000, NAVITAR). Following preliminary measurements, the actuators were submerged in a pressure chamber and subjected to 3000 psi of pressure. Actuators were removed from the chamber and bend angle measurements taken again. Error bars are calculated as standard deviation of the sample set.

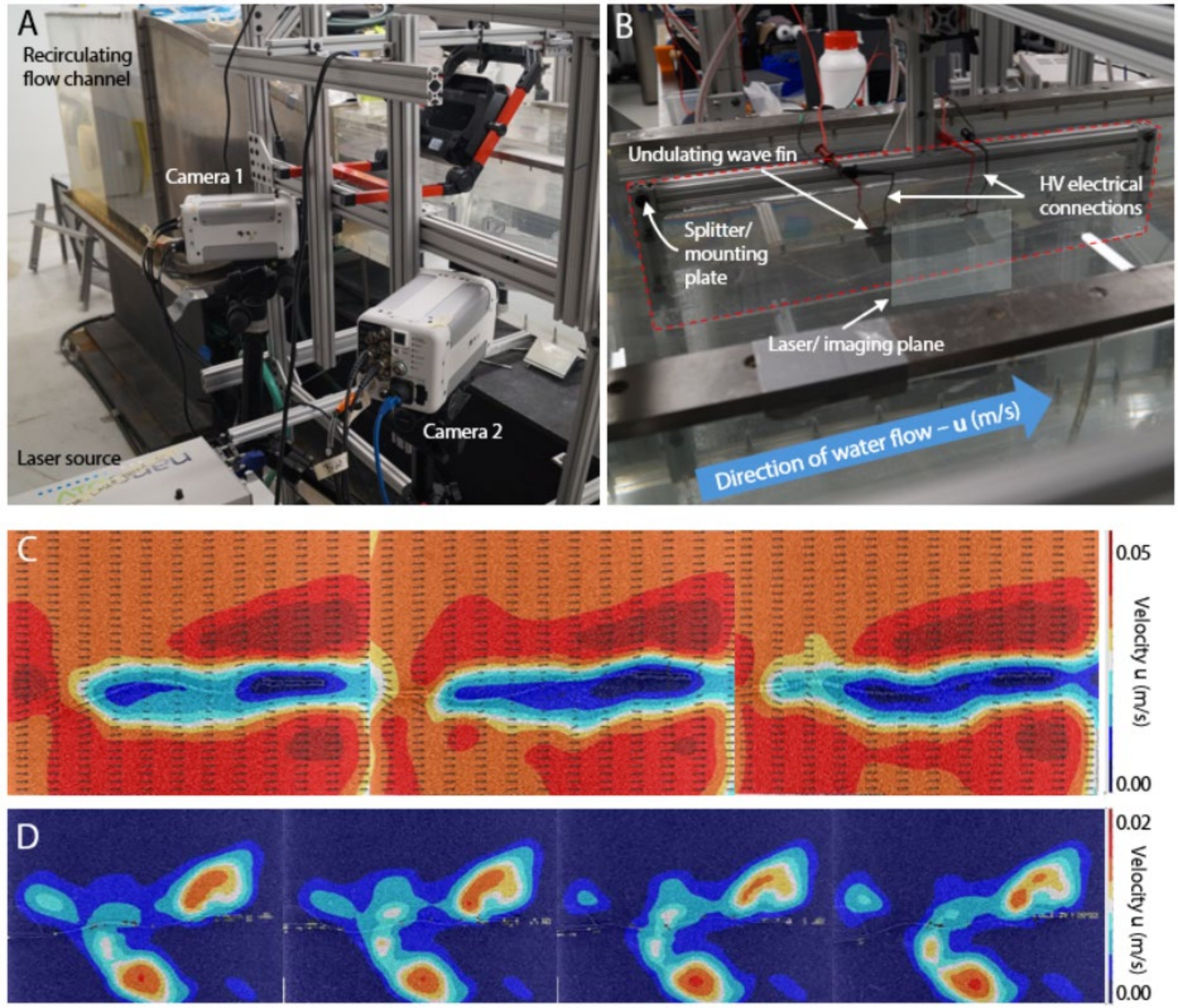

**Fig. S3. Stereo-PIV of the robotic undulating wave fin.** (**A**) Water channel, stereo-camera, and laser setup are shown. Cameras are 9-Megapixel Photron Nova R5 high-speed cameras with LaVision Scheimpflug Adaptors. (**B**) View into the water channel. The robotic fin is mounted to a splitter plate, which can be positioned in space relative to the laser plane within the water channel. Tests were conducted under induced flow (4 cm/s) and no-flow conditions. The direction of water flow and associated vector **u** (m/s) are portrayed within the channel. (**C**) PIV image sequence of a fin performing forward swimming at 4 Hz under 4 cm/s free stream. Images are taken at three equally spaced intervals in a single flapping cycle, highlighting **u**-velocity (in-line). Low velocity is measured at the fin tip due to surface drag, and regions of increased-velocity above and beneath the fin indicate fluid jet formation due to the flapping motion. The results are a phase-average flow field over 20 actuation cycles, with PIV calculated by a single pass of 64x64 pixels with a 50% overlap. (**D**) The image sequence captures **u**-velocity about a fin under 6 Hz actuation forward swimming. Under no-flow conditions, the formation of two primary jets is more apparent, with a third minor jet forming above the leading-edge actuator. The results are a phase-average flow field over 20 actuation cycles, with PIV calculated by a single pass of 64x64 pixels with a 50% overlap.